\documentclass[12pt]{llncs}
\usepackage{multirow}
\usepackage[pdftex,dvipsnames]{xcolor}
\usepackage[]{graphicx}
\usepackage{hyperref}
\usepackage{xargs}
\usepackage{wrapfig}
\usepackage{textcomp}
\usepackage{bm}
\usepackage{amsmath}
\usepackage{gensymb}
\usepackage{booktabs}
\usepackage{subcaption}
\usepackage{booktabs}

\newcommand{\capsnet}{CapsNet}
\newcommand{\capsnets}{CapsNets}

\newcommand{\vcaps}{Vector-CapsNet}
\newcommand{\mcaps}{MatrixEM-CapsNet}

\newcommand{\capss}{vcaps-s}
\newcommand{\capsd}{vcaps-d}

\newcommand{\capst}{vcaps-t}

\newcommand{\emB}{caps-em}

\newcommand{\baseline}{cnn-wp}

\newcommand{\fullwidth}{12.0cm}

\newcommand{\intermediatewidth}{10.0cm}
\newcommand{\fullpath}[1]{./imgs/#1}

\begin{document}
\title{Assessing Capsule Networks With Biased Data}

\titlerunning{Assessing Capsule Networks With Biased Data}
\author{Bruno Ferrarini\inst{1}\and
Shoaib Ehsan\inst{1} \and
Adrien Bartoli\inst{2}\and
Ale\v{s} Leonardis\inst{3} \and
Klaus D. McDonald-Maier\inst{1}
}
\authorrunning{B. Ferrarini et al.}
%
\institute{University of Essex, CSEE, Wivenhoe Park, Colchester CO4 3SQ, UK \\
\email{\{bferra,sehsan,kdm\}@essex.ac.uk} \and
Facult\'{e} de M\'{e}decine, 28 Place Henri Dunant, 63000 Clermont-Ferrand, France \\ 
\email{Adrien.Bartoli@gmail.com} \and
University of Birmingham, School of Computer Science, Birmingham B15 2TT, UK\\
\email{a.leonardis@cs.bham.ac.uk}
}
\maketitle              
\begin{abstract}
Machine learning based methods achieves impressive results in object classification and detection. Utilizing representative data of the visual world during the training phase is crucial to achieve good performance with such data driven approaches. However, it not always possible to access bias-free datasets thus, robustness to biased data is a desirable property for a learning system. Capsule Networks have been introduced recently and their tolerance to biased data has received little attention. This paper aims to fill this gap and proposes two experimental scenarios to assess the tolerance to imbalanced training data and to determine the generalization performance of a model with unfamiliar affine transformations of the images. This paper assesses dynamic routing and EM routing based Capsule Networks and proposes a comparison with Convolutional Neural Networks in the two tested scenarios. The presented results provide new insights into the behaviour of capsule networks.

\end{abstract}
\section{Introduction}
\label{sec:intro}
A robust classification system is expected to give the same prediction for every image of the same class or for images representing the same element in different poses. Machine learning methods, such as Convolutional Neural Networks (CNN), have been used in many classification, detection and recognition tasks \cite{krizhevsky2012imagenet,kalliatakis2017evaluating,arandjelovic2016netvlad}. However, in order to achieve good performance with data driven approaches, well representative data of the visual word are required \cite{masko2015impact,kortylewski2017empirically,karianakis2016empirical}. 
While it is possible to mitigate some bias effects with de-biasing techniques \cite{khosla2012undoing} or with data augmentation \cite{quadnet}, it is important to use machine learning approaches with good generalization performance as it contributes to design more robust applications to unseen or underrepresented imaging conditions. This paper focuses on the latter topic and presents a comparison between Convolutional Neural Networks (CNNs) and Capsule Networks (\capsnets{}) \cite{sabour2017dynamic,hinton2018matrix}. The neurons in a \capsnet{} are organized in groups denoted as Capsules \cite{hinton2011transforming}. In contrast to a single neuron, a capsule can learn a specific image entity over a range of viewing conditions such as viewpoint and rotation. With the use of a routing algorithm to interconnect the capsules, a \capsnet{} model would be affine invariant and spatially aware.
 While the behaviour of CNNs with biased data has been extensively investigated \cite{karianakis2016empirical,kortylewski2017empirically,krizhevsky2009learning},  how bias influences \capsnets{}' performance has received little attention so far. 

This paper aims to fill this gap by proposing two experimental scenarios. The first experiment set evaluates a model's classification accuracy with unfamiliar affine transformations. It introduces a capture bias  \cite{torralba2011unbiased} obtained with  training and test data having transformation intensities sampled from different distributions. The second test scenario is to assess the variation of a network's performance when trained with a dataset presenting several overrepresented classes with respect to evenly distributed classes. The results are presented for five network models: three dynamic routing-based \capsnet{} \cite{sabour2017dynamic} with one, two and three capsule layers respectively, an EM-Matrix routing \capsnet{} \cite{hinton2018matrix} and for a CNN, which represents a comparison baseline.

The rest of this paper is organized as follows. Section \ref{sec:work} provides an overview of related work; Section \ref{sec:caps} gives an introduction on capsule networks;   Section \ref{sec:experiment} describes the method and criteria used for the performance evaluation. The results obtained are presented and discussed  in Section \ref{sec:res}. Finally, Section \ref{sec:conclusions} draws conclusions and proposes possible extensions.
\section{Related Work}
\label{sec:work}
The impact of bias on data driven methods have been extensively explored in the literature. A review of various types of bias in machine learning datasets is provided in \cite{glauner2018impact}. The problem of bias in popular datasets dissected by cause is presented in \cite{torralba2011unbiased} and further discussed in \cite{tommasi2017deeper} where several de-biasing methods are compared. The generalization performance of CNNs is assessed with unfamiliar scale factor in \cite{karianakis2016empirical} and with unfamiliar yaw pose and lighting conditions in \cite{kortylewski2017empirically}, utilizing face recognition tasks. 
The analysis of imbalanced data is addressed in \cite{masko2015impact} and \cite{akosa2017predictive}. In \cite{masko2015impact} several imbalanced datasets are built from CIFAR-10 \cite{krizhevsky2009learning} by means of class down and over-sampling and used to assess CNNs. In \cite{akosa2017predictive}, the importance of choosing the suitable performance evaluation metric in the presence of imbalanced classes is discussed. To the best of our knowledge, the only work addressing the generalization problem for \capsnet{}s is \cite{gritsevskiy2018capsule}, which demonstrates that dynamic routing based \capsnet{}s generalize faster than CNNs  when training data is injected with a few examples of an unfamiliar class. Only a few other works analyze this type of \capsnet{} but without considering bias or generalization performance: \cite{xi2017capsule} and \cite{nair2018pushing} only test \capsnet{}s with more complex data than those utilized in the original paper \cite{sabour2017dynamic}. Our paper aims to fill these gaps by proposing an analysis of the generalization performance with unfamiliar affine transformations and imbalanced training data for both the available architectures of CapsNets: dynamic routing \cite{sabour2017dynamic}  (denoted as \vcaps{} from now on) and EM-Matrix routing based \cite{hinton2018matrix} (\mcaps{}).
\section{Capsule Networks}
\label{sec:caps}
A capsule is a group of neurons whose activity is a tensor which can learn to detect a specific entity over a domain of limited range of viewing conditions such as viewpoint, rotation and lighting \cite{hinton2011transforming}. Two Capsule Networks (\capsnets{}) are proposed in \cite{sabour2017dynamic} and \cite{hinton2018matrix} which are characterized by the architecture outlined as follows. 1) An input stage including one or more regular convolution layers; 2) a single Primary Capsule Layer consisting of a convolutional stage whose neurons are grouped into capsules; 3) one or more Capsule Layers, with the last one as network output, and consists of a capsule per class. Every pair of capsule layers (this includes the Primary layer) are fully connected by means of a routing stage. Routing allows a \capsnet{} to learn relationships between entities by directing the output of a capsule to the proper parent capsule located in the next level. For example, a capsule that learned to recognize eyes, will be routed towards the parent capsule for faces but not to a torso capsule.

\capsnets{} from \cite{sabour2017dynamic} and \cite{hinton2018matrix} have significant differences in their capsule architecture and routing algorithm.
The architecture from \cite{sabour2017dynamic} (\vcaps{}) utilizes 1D vector capsules whose length is an hyperparameter. A capsule encodes an entity and its pose like a CNN, deeper capsules encoding higher level entities. The routing stage fully connects two consecutive capsule layers ($L$ and $L+1$), thus the total input of a capsule ($j$) in $L+1$ depends on the output of every capsule in $L$. Dynamic routing between capsules works as follows. The output ($u_i$) of a capsule is multiplied by a transformation matrix $W_{ij}$ to obtain the prediction vector (\^{u$_{i|j}$}). If the prediction vector is similar to the output of the parent capsule $j$, then the routing algorithm concludes that $i$ and $j$ are highly related and assigns a high value to the related coupling coefficient ($c_{ij}$). As the contribution to the total input of $j$ provided by the capsule $i$ is computed as \^{u$_{i|j}  c_{ij}$}, the coupling coefficient expresses how likely capsule $i$ will activate capsule $j$. Furthermore, the capability of learning relationship between entities that characterize \capsnet{}s is due to a transformation matrix $W$ for each capsule pair $i \in L$ and $j \in \L+1$.

The capsules of the network proposed in \cite{hinton2018matrix} (\mcaps{}) consist of a scalar activation ($a$) and a $4\times 4$ pose matrix ($M$). As in \vcaps{},  capsule layers are fully connected. Thus, each capsule $i$ in a layer $L$ is connected to each capsule $j$ in the next layer $L+1$ by means of a $4\times 4$ transformation matrix ($W_{ij}$) which is learned with an iterative routing algorithm based on EM (Expectation Maximization) clustering and denoted as EM Routing. The prediction of the parent capsule's pose matrix $V_{ij}$ (vote) is computed as the product between $M_i$ and $W_{ij}$ and utilized along with $a_i$ by a routing algorithm to assign routes between capsule $i$ in layer $L$ and capsule $j$ in layer $L+1$ ($\forall i,j$).

The main difference between \capsnet{} and CNN is how features are routed between layers. CNN utilizes single neurons for representing image features and  pooling operations as routing mechanisms. Pooling ensures invariance to small image changes (translation in particular)  at the cost of information loss \cite{lecun2015deep} and makes nearly impossible for a CNN to learn relationship between image entities.
\section{Experimental Setup}
\label{sec:experiment}
The proposed approach consists of two types of experiment to assess a network's performance with unseen affine transformations and with prominent class imbalance. 

\subsection{Capture Bias Experiment}
Training data and test data are built from the same dataset by applying affine transformations whose intensity is sampled from different distributions. Hence, a model becomes familiar with several image transformations which appear at different intensities in the training and test datasets. For example, if the considered transformation is rotation, the training set would be augmented by a rotation angle sampled in a range, such as $[-20^\circ, +20^\circ]$, while the transformation magnitude for testing would be sampled from a wider range such as $[-90^\circ, +90^\circ]$.\\
The performance metric utilized for these experiments is classification accuracy, which is the number of correct predictions from all predictions made. Hence, more general models are those achieving higher accuracy on unseen magnitude of a given affine transformation.

In order to provide more comprehensive insights about the influence of unseen imaging conditions, two different 
criteria for sampling training data are used: uniform and sparse sampling. 
\subsubsection{Uniform Sampling}
\label{sssec:uniform} 
Let $T$ be an affine transformation, $D_r$ a training dataset, $D_e$ the relative test dataset, $R_r$ and $R_e$ two magnitude ranges such that $R_r \subset R_e$. A network is trained with $D_r$ whose every sample $s$, is augmented with $T(s,t_r)$ where $t_r$ is the magnitude uniformly sampled from $R_r$: $t_r \in R_r$. Our tests consist of running the model along the complete axis of transformation range $R_e$. Thus, a set of magnitudes are sampled at fixed size steps starting from the lower bound of $R_e$ until the end of the range. For each $t_{e_i} \in R_e$, the complete dataset $D_e$ is transformed with $T(t_{e_i})$ and used to compute a network's accuracy. This process results in a curve showing the relationship between transformation magnitude and a model's accuracy.\\
\subsubsection{Sparse Sampling}
Let $T$ be an affine transformation, $D_r$ a training dataset, $D_e$ the relative test dataset, $R_r$ and $R_e$ two magnitude ranges such that $R_r \subset R_e$. A subset of $n$ of values are chosen from $R_r$ to form a set $K = \{t_{r_1}, t_{r_2}, \dots t_{r_n}  \}$.  A network is trained with $D_r$ whose sample $s$ is augmented with $T(s,t_r)$ where $t_r$ is the magnitude uniformly sampled from $K$: $t_r \in R_r$. 
Our test procedure is the same as in the Uniform Sampling experiment.
\subsection{Imbalanced Data}
A model trained with imbalanced classes presents a bias towards the overrepresented ones, which results in more frequent prediction of such majority classes \cite{glauner2018impact}. The performance measure is the Matthew’s Correlation Coefficient (MCC) for multiple classes \cite{jurman2010unifying} as it is proven to be more insensitive to imbalanced data than accuracy \cite{akosa2017predictive}. MCC value can fall in $[-1,+1]$, where $+1$ corresponds to a perfect classification. A network is trained with both balanced and imbalanced data and the resulting MCC values are compared. Better models are expected to have a narrower gap between MCC scores of balanced and imbalanced data.\\ 
\section{Results}
\label{sec:res}
\begin{center}
\begin{table}[b!]
\caption{Models assessed: \baseline{} is a CNN similar to the comparison baseline from \cite{sabour2017dynamic}, \capss{}, \capsd{} and \capst{} are \vcaps{} with single, double and triple capsule layers respectively, \emB{} is a \mcaps.}
\begin{tabular}{|@{\hspace{1em}}l@{\hspace{1em}}|@{\hspace{1em}}l@{\hspace{2em}}|}
\toprule
Model	&	Layers \\  \midrule
\baseline{} 	&	$\text{C}(5,1,256); \text{P}(3,1); 2\times[\text{C}(5,1,256); \text{P}(3,2)]; \text{F}(328,192,10)$	\\
\capss{}		&	$\text{C}(9,1,256);\text{C}(9,2,256); \text{Pr}(1152,8,3); \text{Cps}(10,16)$\\
\capsd{}		&	$\text{C}(9,1,64);\text{C}(9,2,64); \text{Pr}(288,8,3); \text{Cps}(20,10); \text{Cps}(10,16)$		\\
\capst{}		&    $\text{C}(9,1,128);\text{C}(9,2,128); \text{Pr}(1152,4,3); 2\times[\text{Cps}(32,8)]; \text{Cps}(10,16)$\\
\emB{}		&	$\text{C}(6,2,32) ; \text{A:} 32; \text{B:} 24; \text{C:} 24; \text{D:} 24; \text{E:} 10; \text{K:} 3 ;$\\ \bottomrule
%
\end{tabular}
\label{tab:models}
\end{table}
\end{center}
Results are presented for several models as listed in Table \ref{tab:models}: \baseline{} is a
CNN with three layers and max pooling, \capss{}, \capsd{} and \capst{} are \vcaps{} with one, two and three layers of capsules respectively and \emB{} is a \mcaps{}. All the networks are implemented with Tensorflow \cite{tensorflow2015-whitepaper}.
In particular, \capss{}, \capsd{} and \capst{} are built on top of the source code provided by the authors  of \vcaps{} \cite{vcaps_tf}, while \emB{} is derived from the code shared at \cite{matrixEM_tf}.
The \baseline{} model is implemented from scratch and has similar architecture and hyperparameters as the comparison baseline from \cite{sabour2017dynamic} used to evaluate \vcaps{} on the MNIST dataset \cite{lecun1998gradient}.
For the notation in Table \ref{tab:models}, the following convention is utilized. C$(k,s,o)$ represents a convolutional layer with kernel $k$, stride $s$ and $o$ filters; P$(k,s)$ indicates a max pool layer with kernel $k$ and stride $s$;   F$(i, h, o)$ is a fully connected network with a single hidden layer of $h$ neurons; Pr$(c,l,r_i)$ indicates a Primary Capsule Layer having $c$ capsules with length $l$ and utilizing $r$ iterations for the routing algorithm; Cps$(c,l,r)$ represents a capsule layer and $c$, $l$ and $r$ have the same meaning as for Pr$(c,l,r)$. Except for an additional convolutional layer at the start, \emB{} has the same architecture as proposed in \cite{hinton2018matrix} but uses less capsules per layer. While in \cite{hinton2018matrix} the hyperparameters $A$, $B$, $C$, $D$ are all equal to 32, our implementation reduces the complexity of the network by setting B, C and D to $24$. This compromise was necessary to run \emB{} with at least 2 routing iterations on our 8GB RAM graphics card.
The models have been trained with the Adam \cite{kingma2014adam} optimizer with default parameters ($\beta_1 = 0.9$ and $\beta_2 = 0.999$) and with an initial learning rate of $0.001$ for \vcaps{} and \baseline{}, and $0.0005$ for \mcaps{}. The loss function to train \capss{}, \capsd{}  and \capst{} is Margin Loss \cite{sabour2017dynamic}  with parameters ${m^{+}=0.9, m^{-}, \lambda=0.5}$. The Spread Loss \cite{hinton2018matrix} has been used for \emB{} with margin $m$ increasing from $0.2$ up to $0.95$ in around 10 epochs. Regularization has been obtained with a reconstruction stage consisting of a neural network with two hidden layers of $512$ and $1024$ units respectively. 
\subsection{Generalization Performance on Unfamiliar Affine Transformations}
\label{ssec:aff_results}
\begin{figure}[t!]
\centering
\includegraphics[width=\intermediatewidth{}]{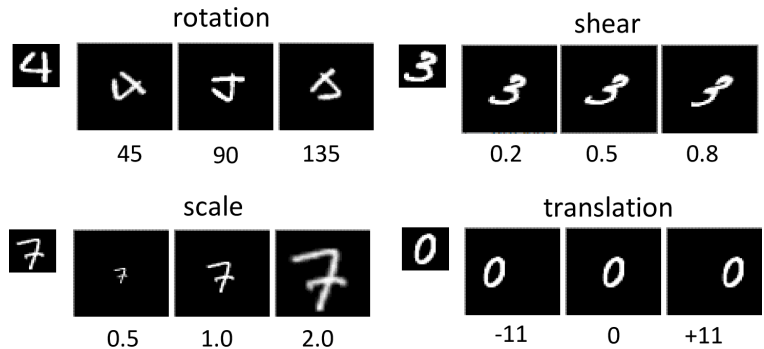}
\caption{Several MNIST images as they are transformed and padded for testing a model accuracy.}
\label{img:MNIST}
\end{figure}
Generalization performance with uniformly sampled affine transformations (Section \ref{sssec:uniform}) has been assessed utilizing affMNIST \cite{affMNIST} as training data and MNIST \cite{lecun1998gradient} for tests. AffMNIST is a dataset  obtained from MNIST by applying to each image several uniformly sampled transformations, namely rotation in $[-20^{\circ}, 20^{\circ}]$, scale between $0.8$ and $1.2$, shear along the $x$ axis in $[-0.2, 0.2]$ and translation. As compared to MNIST, which has 28 pixel images, affMNIST has 40 pixel images in order to fit scaled up digits.  Accuracy data is obtained for each transformation using the MNIST test set with the following extended ranges: rotation $[-90^{\circ},90^{\circ}]$, scale factor $[0.5, 2.0]$,  horizontal shear $[-0.8,0.8]$ and horizontal translation ($x$ axis) $[-13, 13]$. As test required wider range of transformations with respect to those available during training, the models have been fed with  56 pixel images obtained by zero-padding affMNIST images. 
Padding allowed us to test the models with scale factors up to 2.0 and wider translations than those present in affMNIST without any crop to MNIST digits. Figure \ref{img:MNIST} shows some samples from MNIST as they are transformed and padded for testing a model accuracy.

\begin{figure}[ht!]
\centering
\includegraphics[width=\fullwidth{}]{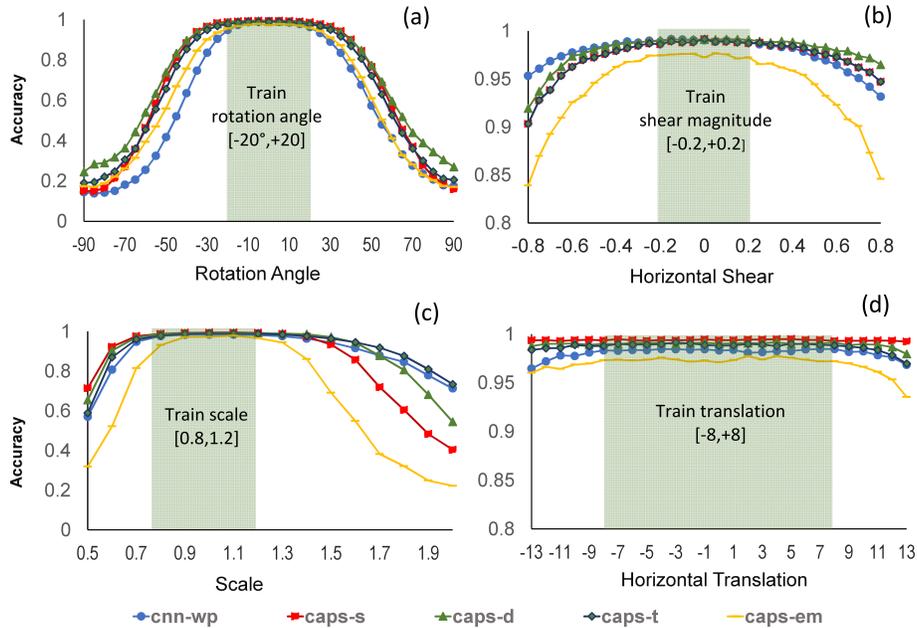}
\caption{Accuracy as a function of Rotation Angle (a), shear along the $x$ axis (b), scale factor (c)  and horizontal Translation (d). The green area indicates the affine transformation range available in training data (affMNIST).}
\label{img:plots}
\end{figure}

The results for uniform sampling experiments are shown in Figure \ref{img:plots} where the accuracy as a function of an affine transformation is plotted for each model.
The most prominent difference among models occurs with unfamiliar scales where \capst{} outperforms both \baseline{} and the other capsule networks. A closer look at the scale plot  (Figure \ref{img:plots}.c) allows us to infer a positive relationship between the number of capsule layers in \vcaps{}s and generalization performance with unfamiliar scale factors. Indeed, \capst{} achieves better  accuracy at each unfamiliar scale than \capss{} and \capsd{} for scale factors larger than 1.2, which is the largest scale present in affMNIST. On the contrary, for small test scale this trend is inverted and it appears that \vcaps{} has the slowest decay in accuracy among the considered models.  Also with rotation, \capsnets{} generalize better than other types of networks, keeping the accuracy above $90\%$ in the interval $[-35^{\circ}, 35^{\circ}]$, which is $15^{\circ}$ wider than the sample interval for the rotation used to generate affMNIST.

The same four affine transformations have been considered in sparse sampling experiments. Model training is carried out by augmenting MNIST samples with a single transformation a time whose intensity is sampled in a finite set. Hence, rotation is sampled in $\{-90^{\circ}, 0, +90^{\circ}\}$ , scale in $\{0.5, 2.0\}$,  horizontal shear in $\{-0.5,0.5\}$ and horizontal translation in $\{-11, 11\}$.

\begin{figure}[t!]
\centering
\includegraphics[width=\fullwidth{}]{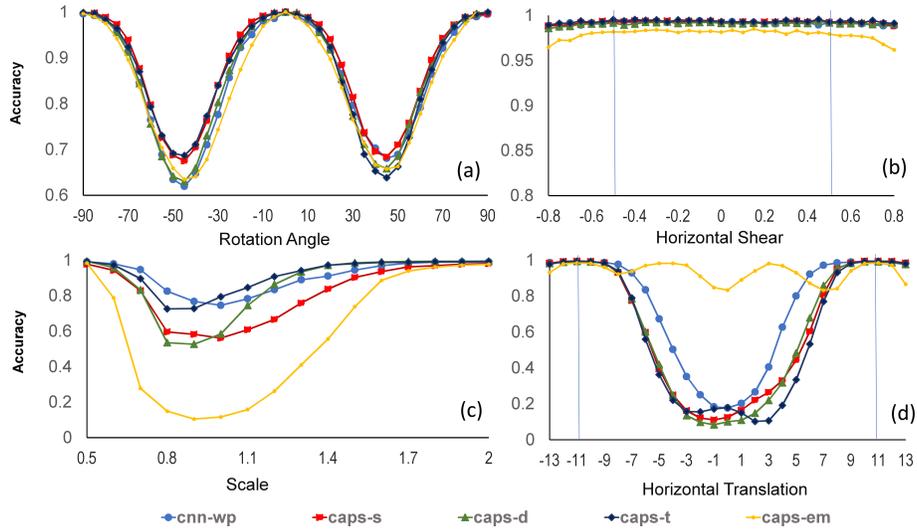}
\caption{Effect of sparse sampling of affine transformations in the training data. Accuracy is represented as a function of Rotation Angle (a), shear along the $x$ axis (b), scale factor (c)  and horizontal Translation (d).}
\label{img:sparse}
\end{figure}

The models do not present significant differences with respect to each other for rotation and horizontal shear (Figure \ref{img:sparse}). In particular, the networks show a very good generalization performance to unseen shear magnitudes. In fact, just including two values for shear in the training set, yields an almost flat accuracy plot along all shear test range. 
Generalization performance with sparse shear sampling is coherent with the results obtained with uniform sampling. Indeed, the models' accuracy has a flat trend along the entire test interval $\{-0.8, 0.8\}$. 
Similarly to the uniform sampling scenario, the scale results show that deeper \vcaps{}s generalize better than the other models with unfamiliar scale factors.

The results from sparse translation experiments show that \baseline{} and the three considered \vcaps{} have a prominent accuracy drop in the middle of the test interval, while \emB{} has stable accuracy on the entire test interval. The reason for the performance gap between \emB{} and \vcaps{} is probably due to the routing algorithm, which is the main difference between these two types of network (Section \ref{sec:caps}).
\subsection{Performance Analysis with Imbalanced Data}
\label{ssec:imba}
\begin{figure}[t!]
\centering
\includegraphics[width=\fullwidth{}]{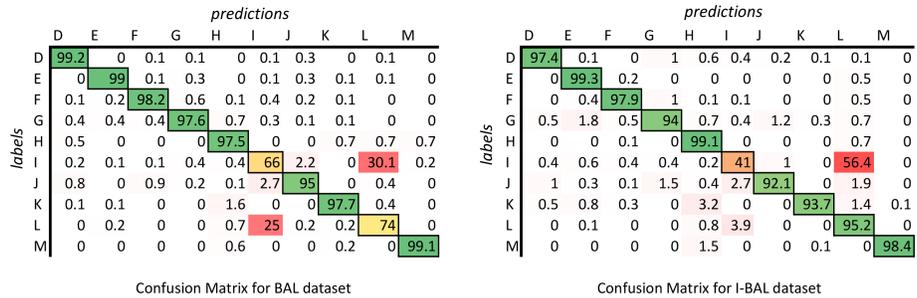}
\caption{Confusion matrices of the \capst{} model for BAL and I-BAL. The over-represented classes $E$, $H$ and $L$ are more often predicted by the model trained with I-BAL thus, this results in misclassification increase.}
\label{img:CM}
\end{figure}
\begin{table}[b!]
\caption{The models' accuracy with MNIST and affMNIST and the models' MCC with balanced (BAL) and imbalanced (I-BAL) datasets. GAP shows the difference between BAL and I-BAL MCC values.}
\label{tab:results}
\begin{center}
\begin{tabular}{|@{\hspace{1em}}l@{\hspace{1em}}|@{\hspace{1em}}c@{\hspace{3mm}}c@{\hspace{1em}}|@{\hspace{1em}}c@{\hspace{3mm}}c@{\hspace{3mm}}c@{\hspace{1em}}|}

\toprule
model   & MNIST        & affMNIST      & BAL    & I-BAL  & GAP     \\ \midrule
cnn-wp  & 0.9923       & 0.9926        & 0.9258 & 0.9021 & -0.0237 \\
caps-s  & 0.9958       & 0.9999        & 0.9202 & 0.8973 & -0.0229 \\
caps-d  & 0.9935       & 0.9981        & 0.9336 & 0.8929 & -0.0407 \\
caps-t  & 0.9933       & 0.9999        & 0.9139 & 0.9004 & -0.0135 \\
caps-em & 0.9827       & 0.9961        & 0.8899 & 0.7483 & -0.1416 \\ \bottomrule
\end{tabular}
\end{center}
\end{table}
The datasets utilized for these experiments have been generated from EMNIST-Letters \cite{cohen2017emnist}, which consists of 26 balanced classes of handwritten letters with 4800 samples each. The balanced dataset (BAL) is a subset of EMNIST including 10 of its classes ($D$ to $M$) with 2400 samples each, while for the imbalanced dataset (I-BAL)  classes have been down-sampled to 600 images, except for $E$, $H$ and $L$ which have the same 4800 samples from EMNIST-Letters. 
Figure \ref{img:CM} shows the confusion matrices of \capst{} for BAL and I-BAL. As expected, the three overrepresented classes, $E$, $H$ and $L$, are predicted more often. This is particularly evident for classes that are similar to each other such as  $L$ and $I$. Indeed, the similarities between lowercase $L$ letters and uppercase 
$I$ letters result in several misclassifications even with BAL datasets where $I$ is predicted as $L$ in $30.3\%$ cases and $I$ is called $L$ in $25\%$ cases. In I-BAL, $L$ is overrepresented as compared to $I$, which is wrongly classified as $L$ more than half of the time ($56.4\%$). 
MCC for all the models are summarized in Table \ref{tab:results}. The least robust model to imbalanced data is \emB{}, with a gap between BAL and I-BAL of 0.1416.  \baseline{} and \capss{} have similar results while \capst{} capture the best performance with a gap of 0.0135, which is about one half of \capss' gap.

The number of capsule layers alone does not explain the better performance of \capst{} over \capss{}.  Indeed, \capsd{} outperforms the other networks with BAL (MCC of $0.9306$) but it also has  the widest gap with unbalanced data among \vcaps{}: 0.0407. Several double layer architectures were examined other than \capsd{}, but it was neither possible to find a better model nor to precisely determine the factor that influences the performance the most. For example, replacing the two capsule layers of \capsd{} (Table \ref{tab:models}) with $2 \times \text{Cps}(128,4,3)$ increased the learnable parameters from $5M$ to $8.9M$ however, the performance decreased sightly from $0.938$ for BAL to  $0.8938$ for I-BAL (with a gap of $0.0442$) in our experiments.
\section{Conclusions}
\label{sec:conclusions}
The analysis of capsule networks has received little attention. This paper aimed to provide novel insights into this new type of neural network and proposed several experiments to assess the performance of a network with biased data. Overall, \capsnet{} outperforms CNNs in most of the cases but not by a large gap. Our results have allowed us to infer that the number of capsule layers (depth) influences generalization performance, this is particularly evident in scale plots (Figure \ref{img:plots}.c) where the accuracy at unseen scales improves with a network depth. Apart from this, the influence of a \capsnet{}'s hyperparameters is not totally understood and would deserve a more detailed and specific analysis. On imbalanced data \capst{} outperforms all the other networks by a consistent gap but the contribution of the triple capsule layer of \capsd{} remains unclear, which is affected by imbalance data more than \capss{}. Finally, the worst model in any scenario is \emB{} with the exception of sparse translation (Figure \ref{img:sparse}). However, it is worth mentioning that the \emB{} implementation it not from its authors and includes less capsules than the model originally proposed in \cite{hinton2018matrix}. Indeed, our Tensorflow implementation is very demanding in terms of RAM and \emB{} is the most complex model that can fit in an 8GB Graphics card.
A natural extension of this work would include \mcaps{} once an official implementation is available. Furthermore, new insights would be provided from a more specific analysis of the relationship between hyperparameters and generalization performance such as the depth and the distribution of capsules among a \capsnet's layers.

\section*{Acknowledgment}

This work has been supported by the UK Engineering and Physical Sciences Research Council EPSRC [EP/K004638/1, EP/R02572X/1 and EP/P017487/1]
%
%
%

\end{document}